\documentclass[twocolumn,numbers,final]{svjour3}

\usepackage{graphicx}
\usepackage{amsmath,amssymb} % define this before the line numbering.
\usepackage{color}

\usepackage{times}
\usepackage{epsfig}
\usepackage{subfigure}
\usepackage{array}
\usepackage{multirow}
\usepackage{natbib}
\usepackage{dsfont}

\newcolumntype{V}{>{$\vcenter\bgroup\hbox\bgroup}c<{\egroup\egroup$}}
\def\Hline{\noalign{\hrule height 4\arrayrulewidth}}

\graphicspath{{imgs/}}

\begin{document}

\title{Discriminative Extended Canonical Correlation Analysis for Pattern Set Matching}

\author{Ognjen Arandjelovi\'c}

\institute{Centre for Pattern Recognition and Data Analytics (PRaDA)\\Deaking University\\Geelong 3216, VIC, Australia\\E-mail: \texttt{ognjen.arandjelovic@gmail.com}\\Tel: +61 (0)3 522-73079}

\maketitle

\begin{abstract}
In this paper we address the problem of matching sets of vectors embedded in the same input space. We propose an approach which is motivated by \emph{canonical correlation analysis} (CCA), a statistical technique which has proven successful in a wide variety of pattern recognition problems. Like CCA when applied to the matching of sets, our \emph{extended canonical correlation analysis} (E-CCA) aims to extract the most similar modes of variability within two sets. Our first major contribution is the formulation of a principled framework for robust inference of such modes from data in the presence of uncertainty associated with noise and sampling randomness. E-CCA retains the efficiency and closed form computability of CCA, but unlike it, does not possess free parameters which cannot be inferred directly from data (inherent data dimensionality, and the number of canonical correlations used for set similarity computation). Our second major contribution is to show that in contrast to CCA, E-CCA is readily adapted to match sets in a discriminative learning scheme which we call \emph{discriminative extended canonical correlation analysis} (DE-CCA). Theoretical contributions of this paper are followed by an empirical evaluation of its premises on the task of face recognition from sets of rasterized appearance images. The results demonstrate that our approach, E-CCA, already outperforms both CCA and its quasi-discriminative counterpart constrained CCA (C-CCA), \emph{for all values of their free parameters}. An even greater improvement is achieved with the discriminative variant, DE-CCA.\\~\\
\keywords{Set \and Matching \and Vectors \and Principal \and Angles}
\end{abstract}

\section{Introduction}\label{s:intro}
Central to any applied problem of pattern recognition is the issue of how the entities of interest should be represented. A numerical description based on readily measurable quantities is sought, one which (as much as possible) minimizes variability due to confounding factors and maximizes that due to differing class memberships. Clearly, this is a highly domain specific task. In computer vision, for example, photometric or geometric models may be used to normalize for illumination and viewpoint changes, or to recover the three-dimensional structure of the scene. After this explicit separation of relevant and confounding variables is performed, the problem ultimately becomes that of inferring class boundaries by matching patterns. In this paper we are specifically interested in set-to-set matching, that is, the case when multiple examples from each class are available both for training, and querying using unlabelled input.

\paragraph*{Assembly Approaches.}
The preferred approach to set comparison is inherently governed by the nature of the particular task to which it is applied. Thus, a large number of different set similarity measures have been proposed and successfully used in different recognition problems. Many of these are non-parametric methods, based on comparisons of individual members of sets. The simplest examples included the \textit{minimum minimorum} \cite{Sato2000} and \textit{maximum minimorum} \cite{ViveSudh2007}
(or Hausdorff) distances which reduce inter-set distance to the distance between only a pair of their elements. Others aggregate member similarities over entire sets, or chosen representative subsets \cite{FanYeun2006}.

\paragraph*{Probability Density Approaches.}
Stronger assumptions are made by methods which assume that different data sets corresponding to the same class are drawn from related probability distributions. These may be estimated using non-parametric, semi-parametric \cite{ShakFishDarr2002} or parametric models, and the closeness between them quantified using, amongst many others, the Bhattacharyya \cite{Bhat1943}, Chernoff \cite{Cher1952} and resistor-average \cite{JohnSina2007,AranCipo2006} distances, or an asymmetric similarity measure such as the Kullback-Leibler divergence \cite{KullLeib1951}. A major shortcoming of probability density-based set matching is the underlying premise that statistical properties of a novel, unlabelled set and the corresponding training set are in some sense alike \cite{AranCipo2013}. Implicit in this is the assumption that novel and training data are acquired in similar conditions (such as the viewpoint and illumination) or very robustly normalized -- both are conditions which are difficult to ensure in nearly all cases of interest.

\begin{figure*}[htb]
  \centering
  \footnotesize
  \begin{tabular}{VV}
    \includegraphics[width=0.47\textwidth]{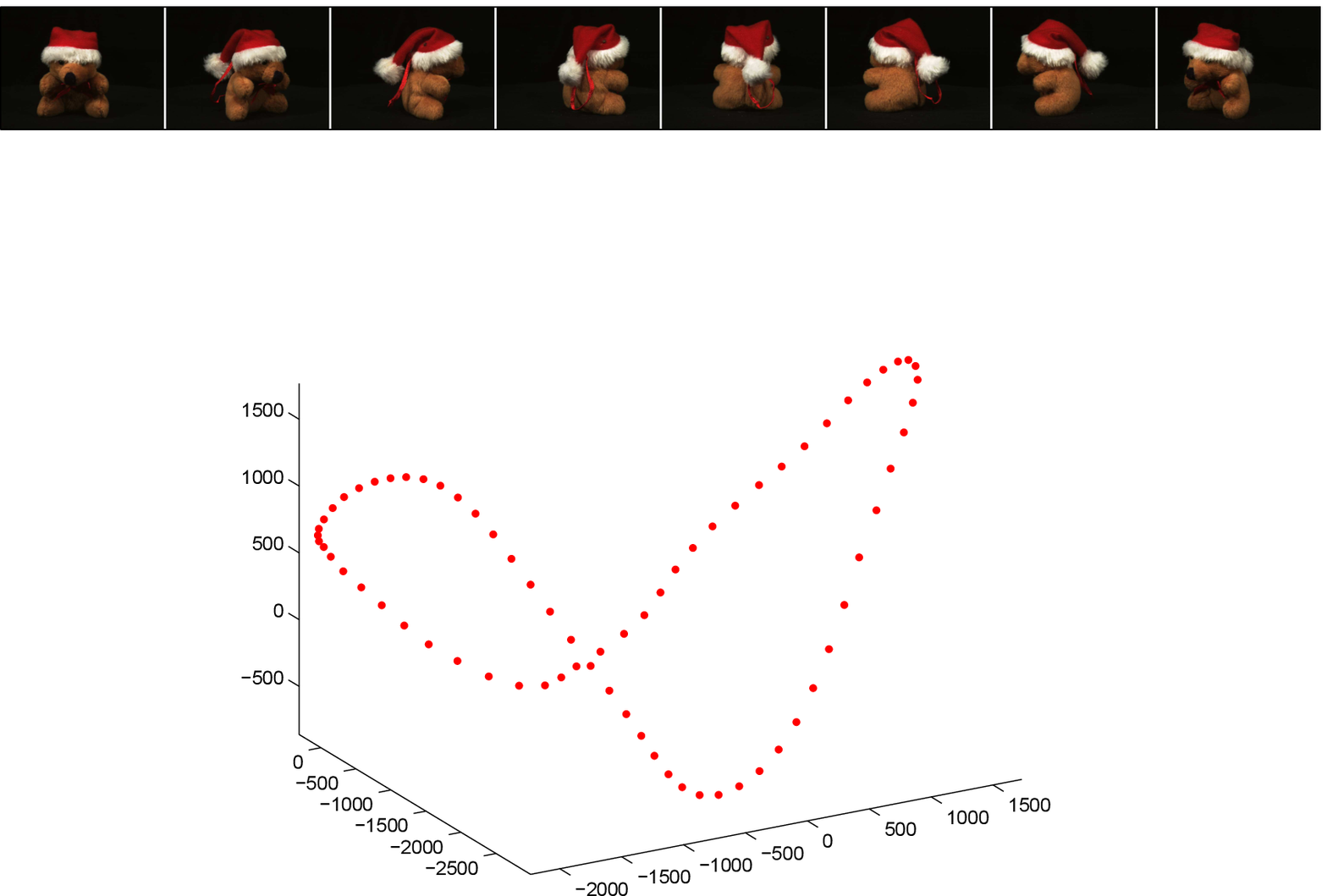} &
    \vspace{15pt}\includegraphics[width=0.47\textwidth]{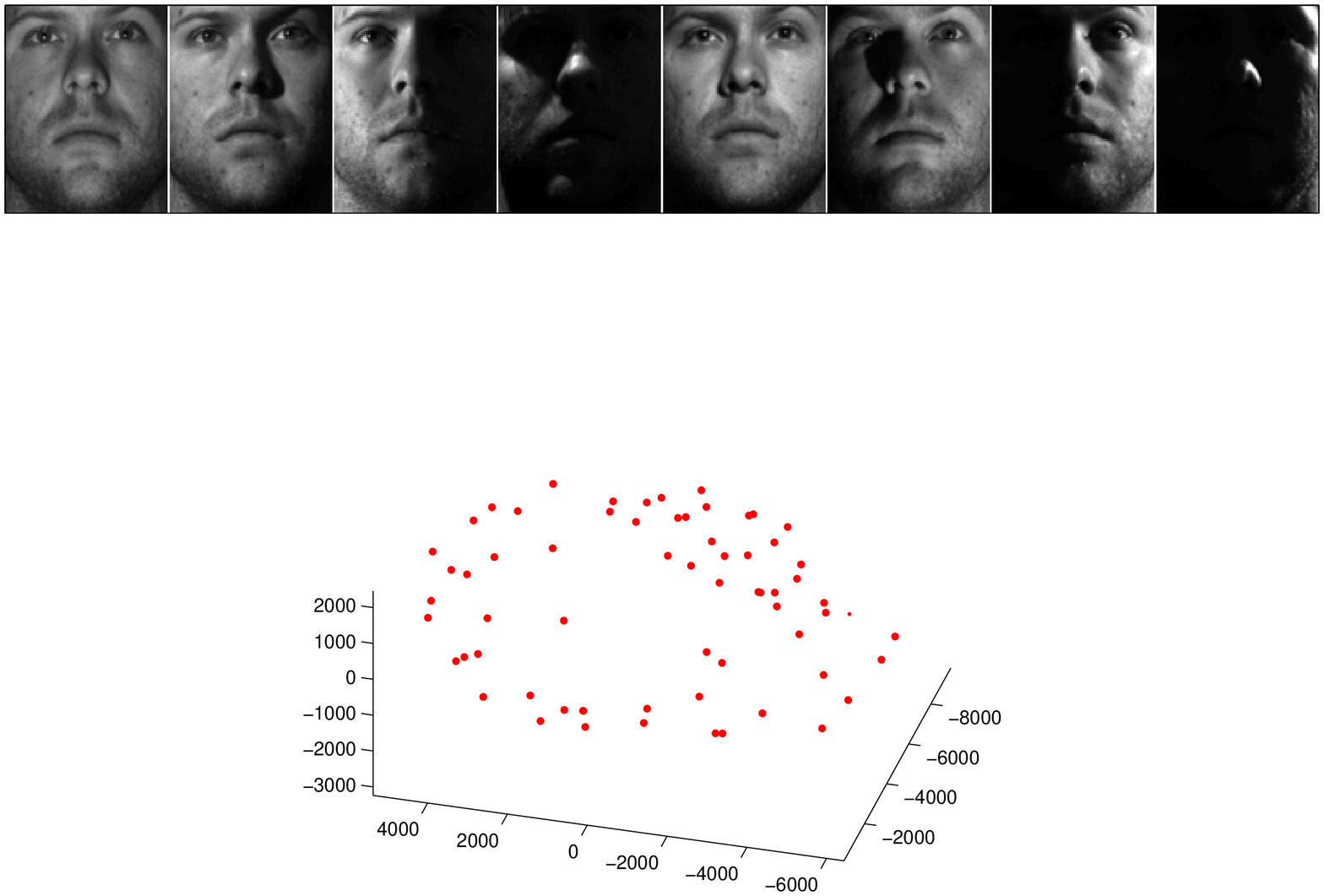}\\
    (a) Object--pose & (b) Face--illumination \\
  \end{tabular}
  \vspace{10pt}
  \caption{ To facilitate learning and be of discriminative use, the representation of patterns must
            possess a certain structure. Often, this structure is exhibited by confining the locus
            of members of each class to a manifold. While their higher order statistics can greatly vary
            (both in the original space, as well as the corresponding class manifold) depending on the
            manner in which data are acquired, the manifold structure can be regarded as containing
            only class-specific information. }
  \label{f:manifolds}
\end{figure*}

\paragraph*{Manifold Approaches.}
While the conditions in which they are acquired can change the observed distribution of data samples, this variation is nonetheless constrained by the intrinsic properties of that class -- often to a manifold, as illustrated in Figure~\ref{f:manifolds}. This can be exploited by learning the structure of this manifold while
disregarding higher order statistics along it \cite{LeeHoYangKrie2003}. A wide range of manifold learning approaches has been described in the literature including multidimensional scaling \cite{BorgGroe2005}, local topology preserving embedding \cite{RoweSaul2001}, eigenspace \cite{GuntBatuAltu+2003} piece-wise linear approximation \cite{LeeHoYangKrie2003,KimAranCipo2007} and nonlinear unfolding using a Mercer kernel \cite{BachJord2002,WolfShas2003,MelzReitBisc2003,Yang2002,FukuBachGret2007}.

The other major issue is that of devising a suitable inter-manifold metric (or rather pseudo-metric). Of particular interest to us are approaches employing \emph{canonical correlation analysis} (CCA), a statistical technique that has been applied with much success to a wide variety of problems, including 3D reconstruction~\cite{ReitDonnLangBisc2006}, infrared to visual image conversion~\cite{DouZhanHaoLi2007}, recognition of texture~\cite{SaisDoreWuSoat2001}, objects~\cite{WolfShas2003,MelzReitBisc2003} and speech~\cite{ChouChol1986}. Broadly speaking, the key idea behind CCA is that a meaningful measure of similarity between two linear (or linearized, as touched upon previously) manifolds can be derived from the most correlated modes of variation between them. Empirical evidence suggests that this indeed is the case in a broad spectrum of problems. What makes CCA additionally attractive in practice is that canonical correlations between linear subspaces can be computed efficiently and in a numerically stable manner \cite{BjorGolu1973}. In this paper we propose a framework which inherits these appealing properties of CCA, whilst at the same time differing from CCA in that when applied to set matching it does not have any parameters which need to be manually tuned and is readily extended to a discriminative framework. Our approach to computing a distance between two sets can be considered manifold-based, but implicitly so, as the distance is computed without explicit manifold fitting. Rather, the distance is robustly inferred by employing second order statistics to account for the confidence that a particular observed intra-set variation corresponds to a phenomenon of interest and not noise. The proposed discriminative framework which follows the main method is based on a similar idea.

These issues and technical details pertaining to CCA and its application to set matching are addressed next, in Section~\ref{s:cca}, followed by Sections~\ref{s:pcca} and~\ref{s:dpcca} introducing respectively extended CCA and discriminative extended CCA, empirical evaluation in Section~\ref{s:eval} and a conclusion with a summary of contributions in Section~\ref{s:conc}.

\section{Set Matching using Canonical Correlation Analysis}\label{s:cca}
Consider two finite sets of vectors, $\mathcal{X} \subset \mathds{R}^{D}$ and $\mathcal{Y} \subset \mathds{R}^{D}$:
\begin{align}
  &\mathcal{X} = \left\{ \mathbf{x}_1, \ldots, \mathbf{x}_N \right\} \\
  &\mathcal{Y} = \left\{ \mathbf{y}_1, \ldots, \mathbf{y}_M \right\}.
  \label{e:sets}
\end{align}
Canonical correlation analysis seeks to find a pair of latent variables or \emph{canonical vectors}~\cite{Hote1936},
$\mathbf{u}_1$ and $\mathbf{v}_1$, such that:
\begin{align}
  &\exists \mathbf{a}_1 \in \mathds{R}^N:~\mathbf{u}_1 = \big[~\mathbf{x}_1 ~|~ \ldots ~|~ \mathbf{x}_N ~\big]~\mathbf{a}_1 = \mathbf{X} \mathbf{a}_1  \label{e:cvectors1}\\
%  \notag \\x
  &\exists \mathbf{b}_1 \in \mathds{R}^M:~\mathbf{v}_1 = \big[~\mathbf{y}_1 ~|~ \ldots ~|~ \mathbf{y}_M ~\big]~\mathbf{b}_1 = \mathbf{Y} \mathbf{b}_1 ,
  \label{e:cvectors2}
\end{align}
which maximizes the \emph{canonical correlation coefficient} $\rho_1 \in [0,1]$ defined as
\begin{align}
  \rho_1 = \max_{\mathbf{u}_1,\mathbf{v}_1}~\frac{\big\langle \mathbf{u}_1,\mathbf{v}_1 \big\rangle}{\| \mathbf{u}_1 \| ~ \| \mathbf{v}_1 \| } = \max_{\mathbf{u}_1,\mathbf{v}_1}~\frac {\mathbf{u}_1^T \mathbf{v}_1} {\| \mathbf{u}_1 \| ~ \| \mathbf{v}_1 \| }.
\end{align}
Canonical vectors and correlation coefficients of higher orders, up to $\min(N,M)$, can be defined recursively
under the constraint of mutual orthogonality between all $\mathbf{u}_i$, as well as all $\mathbf{v}_i$:
\begin{align}
  \forall i,j.~i \neq j:~\big\langle \mathbf{u}_i, \mathbf{u}_j \big\rangle = 0 \text{\hspace{10pt} and \hspace{10pt}}
  \big\langle \mathbf{v}_i, \mathbf{v}_j \big\rangle = 0
\end{align}
By construction, it holds that:
\begin{align}
  1 \geq \rho_1 \geq \rho_2 \geq \ldots \rho_{min(N,M)} \geq 0
\end{align}

\subsection{Set Matching Using CCA}\label{ss:matching}
In most cases the application of CCA to set matching considers sets of vectors over the same type of features \cite{HuaPeib20112,Hott2012,Aran2012b}. For example, each vector may be a rasterized representation of an image as in Section~\ref{s:eval} and \cite{KimAranCipo2007,Aran2012b} for example. The usual manner of applying CCA to set matching consists of three steps. (i) First, an orthonormal basis set $\mathbf{B}_X$ of the subspace characterizing variations within a set is estimated using principal component analysis. Specifically, then $\mathbf{B}_X \in \mathbb{R}^{D \times d_X}$, a matrix with columns consisting of orthonormal basis vectors spanning the $d_X$-dimensional linear subspace embedded in a $D$-dimensional image space, can be computed from the corresponding non-centred covariance matrix $\mathbf{C}_X$:
\begin{align}
  \mathbf{C}_X &= \frac{1}{N}~\sum_{i=1}^N \mathbf{x}_i~{\mathbf{x}_i}^T,
\end{align}
as the row and column space basis of the best rank-$D$ approximation to
$\mathbf{C}_X$:
\renewcommand{\arraystretch}{1}
\begin{align}
  \mathbf{B}_X = \arg \hspace{-12pt}\min_{\scriptsize
  \begin{array}{c}
    \mathbf{B}_X \in \mathbb{R}^{D \times d_X}\\
    {\mathbf{B}_X}^T\mathbf{B}_X = I
    \end{array}} \hspace{-8pt}\min_{\scriptsize
  \begin{array}{c}
    ~~\Lambda \in \mathbb{R}^{d_X \times d_X}\\
    ~~\Lambda_{ij} = 0, i \neq j
    \end{array}}
  {\big\|~\mathbf{C}_X - \mathbf{B}_X~\Lambda~{\mathbf{B}_X}^T~\big\|_F}^2,
\end{align}
where $\| . \|_F$ is the Frobenius norm of a matrix.

The dimensionality $d_X$ of this subspace may be preset, it may be inferred from the distribution of data energy across eigenvector directions or indeed left equal to  $N$ -- the number of data points. (ii) Then, the canonical correlation coefficients $\rho_k$ between two subspaces $\mathbf{X}$ and $\mathbf{Y}$ can be computed using
singular value decomposition (SVD) as the singular values of the matrix ${\mathbf{B}_X}^T \mathbf{B}_Y$ \cite{BjorGolu1973}. (iii) Finally, a similarity measure is computed as a function of canonical correlation coefficients. This is often done by averaging the first (i.e.\ the largest) few \cite{MakiFuku2004}, although more complex learning schemes have been proposed \cite{KimAranCipo2007}.

\subsubsection{Motivation and Advantages.}\label{sss:motiv}
In intuitive terms, canonical vectors extract the most similar modes of variation between two sets, while the corresponding correlation coefficients quantify the degree to which these modes actually match. This focus on that which is common is desirable because it makes the similarity score insensitive to the presence of \emph{differing} modes of variation. These may be present in the data because of different acquisition conditions, or they may indeed correspond to corrupted samples. In contrast, probability density-based methods, such as those using the Bhattacharyya distance or the Kullback-Leibler divergence, do not exhibit such robustness.

\begin{figure}[htb]
  \vspace{20pt}
  \centering
  \includegraphics[width=0.40\textwidth]{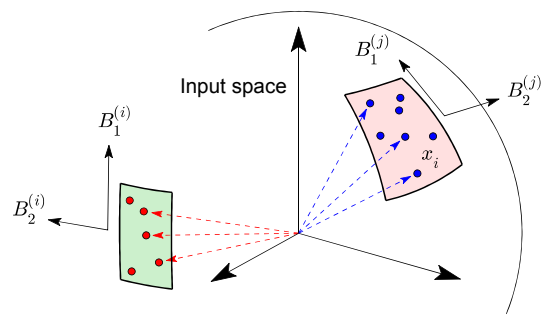}
  \caption{ Canonical correlations are invariant to isotropic scale changes, such as those caused by illumination
            or exposure differences. This effectively constrains the learning of appearance variation to a hypersphere
            centred at the origin. }
  \label{f:hypersphere}
\end{figure}

By modelling variations within a set by a subspace, canonical correlations are also inherently unaffected by uniform scaling (or, equivalently, the contrast) of individual patterns. This makes CCA-based methods particularly suitable for various computer vision tasks, where such variation may be introduced by the changes in
illumination or the duration of exposure of the photosensitive medium. Learning is effectively performed on a hypersphere, as illustrated in Figure~\ref{f:hypersphere}.

Finally, in many cases data variations within a single set are low dimensional making CCA-based matching practically
appealing due to computational efficiency and low storage requirements.

\subsubsection{Limitations of CCA.}\label{ss:limitations}
In modelling class variation, the application of canonical correlation analysis inherently requires the partitioning of the input space into two disjoint subspaces:
the class subspace $\mathbf{B}$ of observed variability, and its complementary subspace $\text{null}(\mathbf{B}^T)$. This hard division causes several undesirable consequences.

\paragraph*{Optimal Parameter Choice and Performance Sensitivity.}
In practice the presence of noise in data is unavoidable and the optimal choice of the dimensionality of the
class subspace can seldom be guaranteed \cite{GouFyfe2009}.  Most commonly, this parameter is determined empirically,
using a training and a validation set \cite{KimAranCipo2007}. Not only through ineffective use of data, this approach
is also unattractive due to its amplification of noise in the non-dominant directions of the class subspace and the
loss of information in the discarded orthogonal directions.

\paragraph*{Second Order Statistics.}
To make matters
worse, any principal direction is either included at an equal footing with all others in the class subspace,
or entirely discarded. As we shall demonstrate, the performance of CCA-based matching is very sensitive to
the choice of this parameter. This is particularly pronounced in cases suffering from limited sample size, when
the number of data points needed for robust estimation for as few as two canonical correlations,
is 40 to 70 times the dimensionality of the class subspace~\cite{BarcStev1975}.

\paragraph*{Discriminative Learning.}
The described CCA-based set distance is not discriminative in nature -- sets are compared in an independent, pair-wise manner, without regard for inter-class and intra-class variability. That this cannot be optimal can be seen easily by noting that depending on the application, the same two sets can be regarded as belonging to either the same or different classes: two sets of face appearance images of two different individuals correspond to the same class if the problem is that of face detection (classification to ``face'' and ``non-face''), and different classes if it is face recognition (classification by the identity).

\begin{figure*}[htb]
  \centering
  \footnotesize
  \begin{tabular}{VV}
    \hspace{10pt} \includegraphics[width=0.3\textwidth]{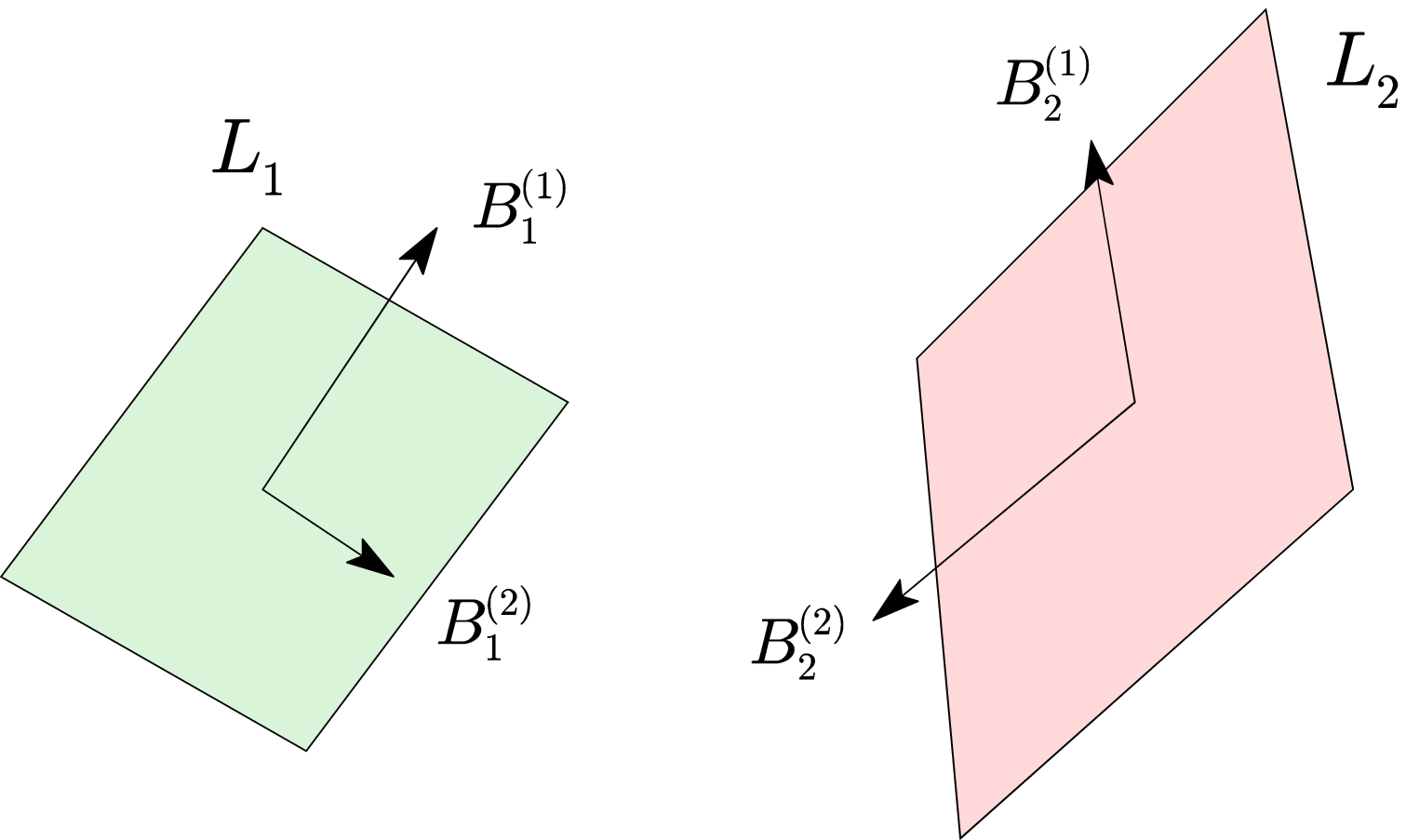} \hspace{10pt} &
    \hspace{10pt} \includegraphics[width=0.4\textwidth]{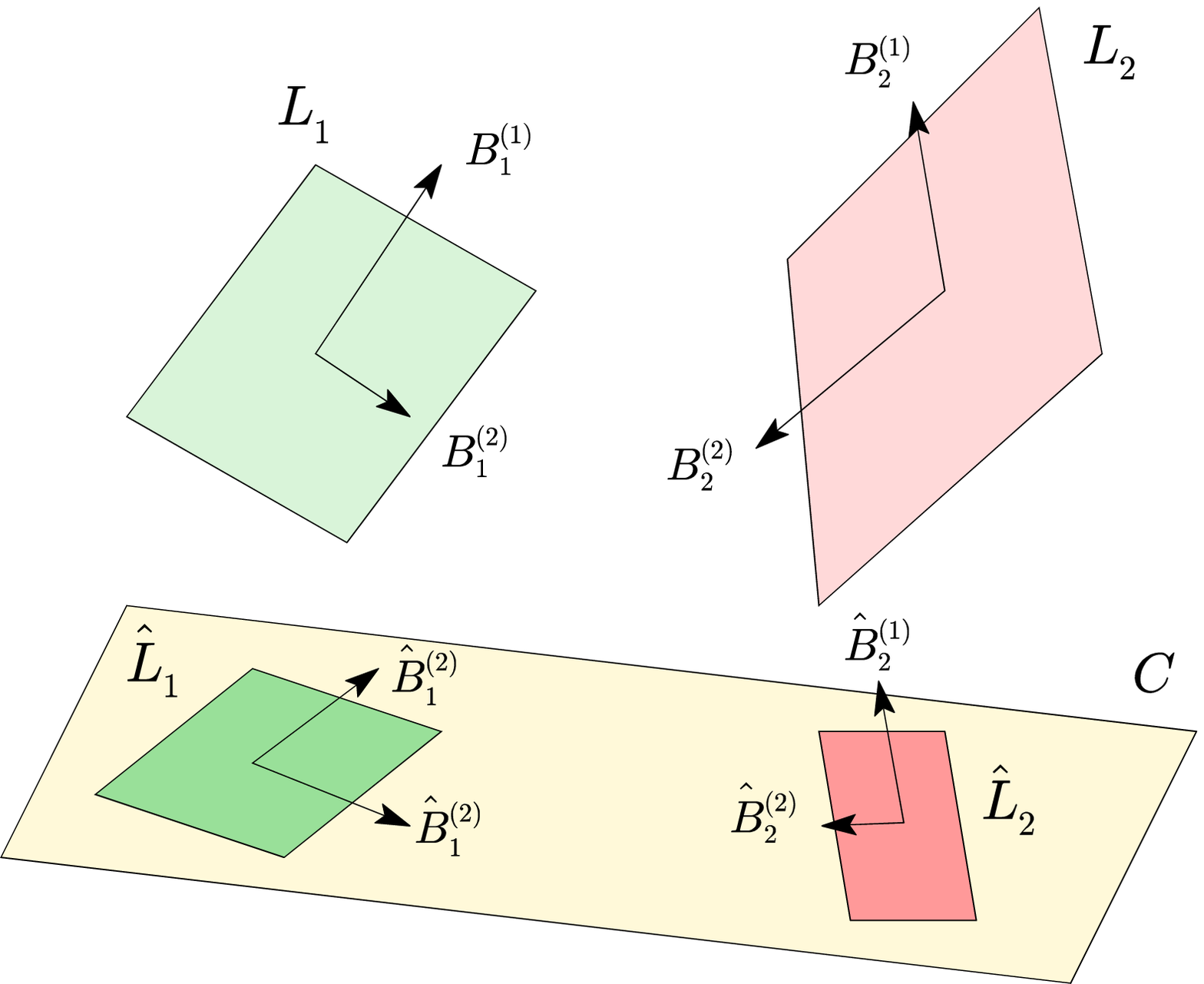} \hspace{10pt}  \\
    (a) CCA & (b) C-CCA\\
  \end{tabular}
  \vspace{10pt}
  \caption{ A conceptual illustration of the principles underlying set matching using
            a similarity measure based on (a) canonical correlation analysis (CCA) and
            (b) its discriminative extension, constrained canonical correlation analysis (C-CCA).
            Both methods effectively compute the angles between subspaces corresponding to
            vector sets, preceded by a quasi-discriminative projection in the case of C-CCA.  }
  \label{f:ccas}
\end{figure*}

The first discriminative extension of CCA was proposed by Oja~\cite{Oja1983}. It consists of a linear projection of data onto a subspace which orthogonalizes basis vectors of different classes. The main shortcoming of this approach lies in its lack of robustness, with the orthogonal discriminative criterion leading to overfitting. A different linear subspace approach, which we will refer to as \emph{constrained canonical correlation analysis} (C-CCA) was described by Fukui \textit{et al.}~\cite{FukuYama2003}. They introduce a discriminative, \emph{constraint} subspace, defined by the principal components corresponding to the \emph{smallest} eigenvalues of the mean projection matrix across all classes:
\begin{align}
  \bar{\mathbf{P}} = \sum_i \mathbf{P}_i= \sum_i \big(\mathbf{B}_i~{\mathbf{B}_i}^T \big)
\end{align}
The computation of canonical correlations is then preceded by a linear projection to the constraint subspace, as illustrated conceptually in Figure~\ref{f:ccas}. For the optimal size of its dimensionality, C-CCA generally outperforms Oja's orthogonal subspace method \cite{KimKittCipo2007}, as well as non-discriminative CCA
\cite{NishYamaFuku2005,AranCipo2006a}. However, ensuring that the optimal value is chosen is difficult. In addition, the construction of the described constraint subspace is \textit{ad hoc} in nature -- it does not maximize any meaningful discriminative function and it does not take into account inter-class variation,
relying purely on intra-class variability. A solution to this problem was proposed by Kim \textit{et al.} \cite{KimKittCipo2007}, in the form of an iterative method which incrementally adjusts the optimal projection subspace to maximize expected inter-class to intra-class canonical correlations. This is achieved at a great computational cost, the loss of closed-form computability and with the restriction of discrimination between only two classes. Also, just like the previous two approaches, this method suffers from an increased number of free parameters and the ``all or nothing'' modelling of class distributions.

\paragraph*{Nonlinearity.} Finally, for completeness, we briefly mention that simple CCA assumes linear intra-class variability. Effective solutions to this problem have been described in the literature, involving either piece-wise linearization of class manifolds \cite{KimAranCipo2007} or their ``unfolding'' \cite{FukuStenYama2006,RoweSaul2001}. All of these approaches eventually reduce to the computation of CCA in its original form and can thus without modification be applied with the methods we propose in this paper.

\section{Extended Canonical Component Analysis (E-CCA)}\label{s:pcca}
Motivated by the standard canonical component analysis, in this section we too seek to find the most correlated modes of variation within two sets. However, unlike before, we wish to do so without discarding any data i.e.\ without the partitioning of the input space into principal (relevant) and complementary (noise) subspaces. As explained in Section~\ref{ss:limitations} this partitioning is a source of major problems when CCA is applied to set matching. For illustration, consider the problem of matching sets of images (see Section~\ref{s:eval} for a practical example). Generally, each of the image pixels is affected by noise which contains a non-correlated component across different pixels. This means that if sufficient data is available, each of the image sets will exhibit variation in all directions of the input image space. Consequently, all of the canonical correlations between any two sets would be $1.0$ if subspace projection described in Section~\ref{ss:matching} was not applied (i.e.\ if the input space was not partitioned). To avoid the need for this projection and the potential for information loss thus effected, we wish to derive a CCA inspired set similarity measure which takes into account the confidence that the observed mode of variation is indeed due to data variability and not noise. We achieve this by incorporating second order statistics into the similarity measure.

Our approach is motivated by observing that for the pair of canonical vectors $\mathbf{u}_1$ and $\mathbf{v}_1$ in Equations~\eqref{e:cvectors1}-\eqref{e:cvectors2} there exists a direction $\mathbf{w}_1$ for which:
\begin{align}
  \mathbf{u}_1 = {\mathbf{B}_X} {\mathbf{B}_X}^T \mathbf{w}_1 \\
  \mathbf{v}_1 = {\mathbf{B}_Y} {\mathbf{B}_Y}^T \mathbf{w}_1
\end{align}
The first multiplication, by ${\mathbf{B}_X}^T$ or ${\mathbf{B}_Y}^T$, has the effect of removing any variation not spanned by the columns of $\mathbf{B}_X$ and $\mathbf{B}_Y$ respectively (the two principal subspaces), while the second multiplication by $\mathbf{B}_X$ or $\mathbf{B}_Y$, re-embeds the vectors in the original input space.

The first canonical correlation coefficient can then be written as:
\begin{align}
  \rho_1 &= \frac {{\mathbf{w}_1}^T~{\mathbf{B}_X} {\mathbf{B}_X}^T \mathbf{B}_Y {\mathbf{B}_Y}^T~\mathbf{w}_1}
        {\|{\mathbf{B}_X}^T\mathbf{w}_1\|~\|{\mathbf{B}_Y}^T\mathbf{w}_1\|} \\
        &=
  \frac {{\mathbf{w}_1}^T~{\mathcal{W}(\hat{\mathbf{\Sigma}}_X)}^T~\mathcal{W}(\hat{\mathbf{\Sigma}}_Y)~\mathbf{w}_1}
        {\|{\mathbf{B}_X}^T\mathbf{w}_1\|~\|{\mathbf{B}_Y}^T\mathbf{w}_1\|}
        \label{e:altCCA}
\end{align}
where $\mathcal{W}(\ldots)$ is the covariance matrix whitening function \cite{DudaHartStor2001}, and:
\begin{align}
   &\hat{\mathbf{\Sigma}}_X = \mathbf{B}_X \hat{\mathbf{\Lambda}}_X {\mathbf{B}_X}^T\\
   &\hat{\mathbf{\Sigma}}_Y = \mathbf{B}_Y \hat{\mathbf{\Lambda}}_Y {\mathbf{B}_Y}^T
   \label{e:proj}
\end{align}

\begin{figure*}[htb]
  \centering
  \footnotesize
  \begin{tabular}{VV}
    \hspace{10pt} \includegraphics[width=0.25\textwidth]{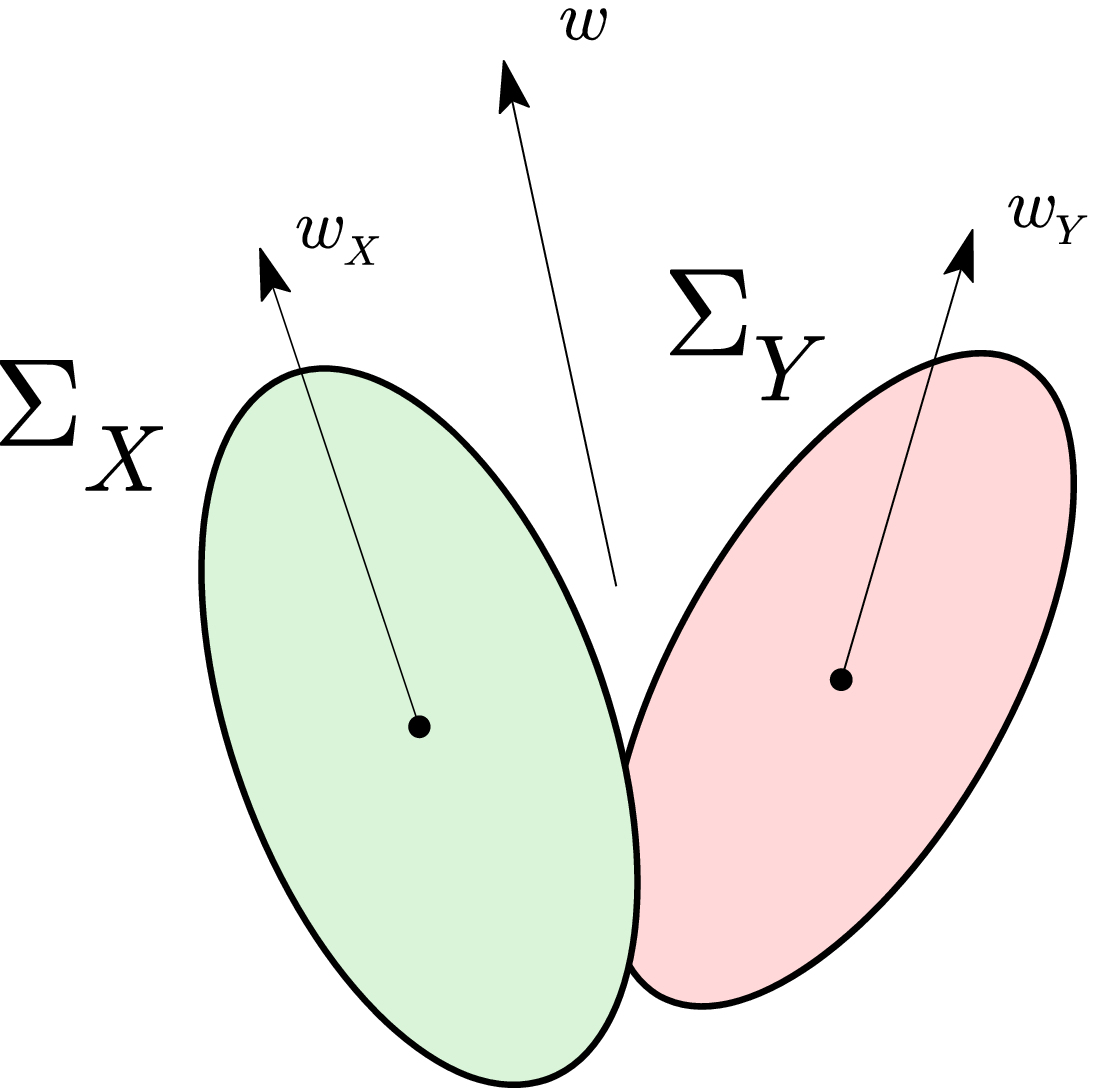} \hspace{10pt} &
    \hspace{10pt} \includegraphics[width=0.45\textwidth]{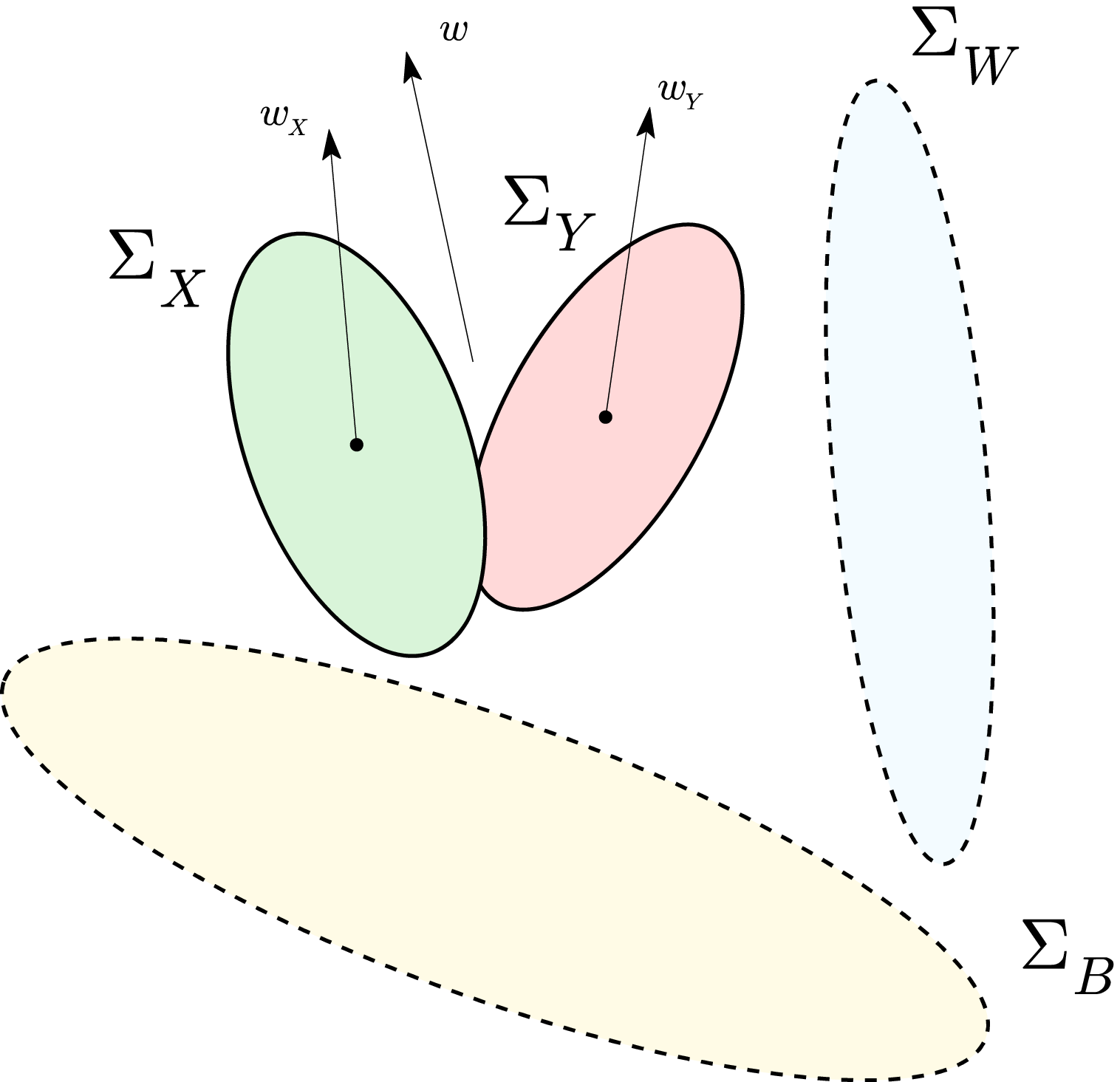} \hspace{10pt}  \\
    (a) E-CCA & (b) DE-CCA\\
  \end{tabular}
  \vspace{10pt}
  \caption{ A conceptual illustration of the principles underlying set matching using
            a similarity measure based on the proposed (a) \emph{extended} canonical
            correlation analysis (E-CCA) and (b) its discriminative variant,
            \emph{discriminative extended} canonical correlation analysis (DE-CCA).
            Set similarity is computed by considering projective space distortions corresponding
            to the sets' covariances and, in the case of DC-CCA, discriminative inter-class
            and intra-class covariances inferred from training data. }
  \label{f:PCCA}
\end{figure*}

Instead of a normalized projection of the vector $\mathbf{w}_1$ onto a subspace, say $\mathbf{B}_X$,
consider its transformation effected by the corresponding ``deviation'' matrix $\mathbf{\Upsilon}_X$ (also
positive semi-definite and symmetric) which we define as:
\renewcommand{\arraystretch}{1.2}
\begin{align}
  \mathbf{\Upsilon}_X &=  \left(\mathbf{\Sigma}_X\right)^{1/2} = {\small\mathbf{V}_X~\begin{bmatrix}
                                           \sqrt{\lambda_X^{(1)}} & 0 & 0 & 0 \\
                                           0 & \sqrt{\lambda_X^{(2)}} & 0 & 0 \\
                                           0 & 0 & \ddots & 0 \\
                                           0 & 0 & 0 & \sqrt{\lambda_X^{(D)}} \\
                                         \end{bmatrix}}~{\mathbf{V}_X}^T \notag\\
                                         &=\mathbf{V}_X~\left(\mathbf{\Lambda}_X\right)^{1/2}~{\mathbf{V}_X}^T
\end{align}
where:
\begin{align}
  \mathbf{\Sigma}_X = \mathbf{X} \mathbf{X}^T = \mathbf{V}_X~\mathbf{\Lambda}_X~{\mathbf{V}_X}^T
\end{align}
is the full data covariance matrix. The said transformation anisotropically scales its input, amplifying it in the directions in which $\mathcal{X}$ exhibits significant variability (large $\lambda_X^{(i)}$) and attenuating in those with little (or indeed no) variability (small $\lambda_X^{(i)}$), as illustrated in Figure~\ref{f:PCCA}. This effect can be considered a generalization of the projection described in Equation~\eqref{e:proj} -- while in the application of standard CCA all variation in the complementary subspace is discarded and the variation in the principal subspace whitened, here the transformation has the effect of smoothly emphasizing or de-emphasizing different directions of variability according to its extent (specifically, proportionally to the standard deviation in the corresponding direction). In the special case of data which exhibits isotropic variability constrained to a subspace the result is exactly the same as in Equation~\eqref{e:proj} (up to scale).

Motivated by this intuition and extending the analogy to Equation~\eqref{e:altCCA}, we seek such unit vector $\hat{\mathbf{w}}_1$ whose projections by $\mathbf{\Upsilon}_X$ and $\mathbf{\Upsilon}_Y$ have the highest degree of correlation. Formally, we define the first extended canonical
correlation coefficient $\psi_1$ between $\mathcal{X}$ and $\mathcal{Y}$ as:
\begin{align}
  \psi_1 &= \max_{\hat{\mathbf{w}}_1}~\big\{(\mathbf{\Upsilon}_X~\hat{\mathbf{w}}_1)^T~(\mathbf{\Upsilon}_Y~\hat{\mathbf{w}}_1) \big\} \\&=
  \max_{\hat{\mathbf{w}}_1}~\big\{{\hat{\mathbf{w}}_1}^T~{\mathbf{\Upsilon}_X}^T~\mathbf{\Upsilon}_Y~\hat{\mathbf{w}}_1 \big\} \\
  &=\max_{\hat{\mathbf{w}}_1}~\big\{{\hat{\mathbf{w}}_1}^T~{\mathbf{\Upsilon}_X}^T~\mathbf{\Upsilon}_Y~\hat{\mathbf{w}}_1 \big\} \\&=
  \max_{\hat{\mathbf{w}}_1}~\big\{{\hat{\mathbf{w}}_1}^T~\mathbf{\Phi}_{XY}~\hat{\mathbf{w}}_1\big\},
  \label{e:pccc}
\end{align}
under the constraint:
\begin{align}
&\| \hat{\mathbf{w}}_1 \| = 1.
\end{align}
The key idea here is that this measure will favour those directions of the space in which both $\mathcal{X}$ and $\mathcal{Y}$ have significant variability (it is ``amplified'' both by $\mathbf{\Upsilon}_X$ and $\mathbf{\Upsilon}_Y$). Similarly, a direction in which $\mathbf{\Upsilon}_X$ (say) exhibits significant variability, but $\mathbf{\Upsilon}_Y$ does not, will contribute to $\hat{\mathbf{w}}_1$ less, while a direction in which neither of the sets exhibit significant variability will be greatly de-emphasized and have little effect on $\hat{\mathbf{w}}_1$.

\label{p:phi}Note that although matrices $\mathbf{\Upsilon}_X$ and $\mathbf{\Upsilon}_Y$ are symmetric, their product
is not. Nonetheless, $\mathbf{\Phi}_{XY}$ is positive semi-definite. This is because linear transformations effected by $\mathbf{\Upsilon}_X$ and $\mathbf{\Upsilon}_Y$
involve no rotation, reflection or shearing, i.e.\ they can be expressed purely as a combination of orthogonal projection,
and (generally anisotropic) scaling. Thus $\psi_1$ is maximized when $\hat{\mathbf{w}}$ is in the direction of the eigenvector
of $\mathbf{\Phi}_{XY}$ corresponding to its largest eigenvalue, for which:
\begin{align}
  &\psi_1 = \lambda_{\Phi}^{(1)},
\end{align}
where $\lambda_{\Phi}^{(1)} \geq \lambda_{\Phi}^{(2)} \geq \ldots \lambda_{\Phi}^{(D)} \geq 0$ are the eigenvalues
of $\mathbf{\Phi}_{XY}$.

\paragraph*{}
When classical canonical correlation analysis is applied to pattern recognition in practice, only the first few correlation coefficients are computed. This is largely a consequence of the trade-off between the accuracy of matching and its speed. Our approach does not suffer from the same weakness. Extending the analysis to higher order extended correlation coefficients, we can quantify the degree of agreement between variations observed in sets $\mathcal{X}$ and $\mathcal{Y}$ as:
\begin{align}
  \mu_{XY} = \frac { \sum_{i=1}^D \psi_i} { \sum_{i=1}^D \sqrt{\lambda_X^{(i)} \lambda_Y^{(i)}} }
  \label{e:pcca_sim}
\end{align}
More elaborate combinations of $\psi_i$ are possible, e.g.\ as described in \cite{KimAranCipo2007}, but here we adopt a simple normalized summation because it lends itself to particularly efficient computation, as we will show shortly.

Note that $\mu_{XY}$ reaches its \textit{maximum maximorum} when $\mathbf{\Upsilon}_X$ and $\mathbf{\Upsilon}_Y$
share the same eigenspace and when their eigenvectors of the same rank (with respect to the magnitude of
the corresponding eigenvalue) are aligned. However, in general:
\begin{align}
  0 \leq \mu_{XY} \leq 1.
\end{align}
The class similarity $\mu_{XY}$ in Equation~\eqref{e:pcca_sim} can be rapidly computed by noticing that:
\begin{align}
  \sum_{i=1}^D \psi_i = \text{Tr}~\big[~\mathbf{\Phi}_{XY}~\big],
\end{align}
while the values of $\lambda_X^{(1)}, \ldots, \lambda_X^{(D)}$ and $\lambda_Y^{(1)}, \ldots, \lambda_Y^{(D)}$
are estimated in the same manner as in the case of classical CCA.

It is important to observe that in Equation~\eqref{e:pccc} there is no concern of an ill-defined result because of a vanishing denominator (unlike in the case of the method described in \cite{AranCipo2006e} for example). Specifically, since $\lambda^{(i)}_X$ and $\lambda^{(i)}_Y$ are ordered in magnitude, the product $\lambda^{(1)}_X~\lambda^{(1)}_Y$ cannot be 0 as both sets $\mathcal{X}$ and $\mathcal{Y}$ are assumed to contain at least some variability. Indeed, this is a necessary condition both for classical CCA and the proposed method to be meaningful in this context.

\section{Discriminative Extended Canonical Component Analysis (DE-CCA)}\label{s:dpcca}
In the previous section we derived a principled extension of canonical correlation analysis suitable for matching sets of patterns constrained to linear subspaces. We addressed the inherent limitations of CCA when applied on this problem: the inevitable ``hard'' partitioning of the input space, as well the practical difficulty of parameter estimation. Unlike classical CCA, we now show that our method readily lends itself to a discriminative learning framework.

In Section~\ref{s:pcca} we considered space transformation by means of projection using the square root of the class covariance matrix:
\begin{align}
  \mathbf{w}_1 \longrightarrow  \mathbf{\Upsilon}_X \mathbf{w}_1 = \left(\mathbf{\Sigma}_X\right)^{1/2} \mathbf{w}_1.
\end{align}
Its  effect is the amplification of modes of variation common to the input vector $\mathbf{w}_1$ and the set $\mathcal{X}$. However, this is achieved without any knowledge of variability both between and within different classes. Here we capture these through two covariance matrices, the \emph{normalized inter-class scatter matrix} $\mathbf{\Sigma}_B$ and the \emph{mean intra-class scatter matrix} $\mathbf{\Sigma}_W$. Using the notation $\mathbf{Z}_i$ for different training set data matrices $i=1,\ldots,N_C$ (each corresponding to a different class) and denoting their members by $\left\{ \mathbf{z}_{i1}, \mathbf{z}_{i2}, \ldots \right\}$,
we define the normalized inter-class scatter matrix as follows:
\begin{align}
   \mathbf{\Sigma}_B =\frac{1}{N_C}~\sum_{i=1}^{N_C} \left( \frac{E[\mathbf{Z}_i]}{\|E[\mathbf{Z}_i]\|} - \mathbf{m} \right)
   \label{e:inter}
\end{align}
where:
\begin{align}
   &\mathbf{m} =\frac{1}{N_C}~\sum_{i=1}^{N_C} \frac{E[\mathbf{Z}_i]}{\|E[\mathbf{Z}_i]\|}.
\end{align}
Note that the explicit normalization of class data means $E[\mathbf{Z}_i]$ in Equation~\eqref{e:inter} is necessary here (see Section~\ref{sss:motiv}). The mean intra-class scatter matrix is simply:
\begin{align}
   \mathbf{\Sigma}_W =\frac{1}{N_C}~\sum_{i=1}^{N_C} {\mathbf{Z}_i}^T \mathbf{Z}_i = \frac{1}{N_C}~\sum_{i=1}^{N_C} {\mathbf{\Sigma}_{Z_i}}
\end{align}
Our definitions of inter-class and intra-class matrices are similar to those used in Fisher's discriminant analysis \cite{DudaHartStor2001}.

The two scatter matrices are then used to further transform an input vector, first by accentuating its components in the direction of common intra-class variations and then by attenuating those corresponding to inter-class variability: \begin{align}
  \mathbf{w}_1 \longrightarrow  \left({\mathbf{\Sigma}_B}\right)^{-1/2}
                                \left({\mathbf{\Sigma}_W}\right)^{1/2} \mathbf{\Upsilon}_X \mathbf{w}_1.
\end{align}
The intuition behind this transformation is exactly the same as that for the non-discriminative version in Section~\ref{s:pcca}. Instead of partitioning the space into discriminative and non-discriminative subspaces and projecting the data onto the former, like in CCA for example, our transformation of data is smoother in nature. While a projection onto the discriminative subspace entirely aligns the data with the subspace, our transformation instead realigns the data smoothly first by emphasizing discriminative directions (according to the inter-class scatter matrix) and then by further de-emphasizing the non-discriminative ones (according to the intra-class scatter matrix). The idea is illustrated conceptually in Figure~\ref{f:PCCA}(b). Note that the aforementioned directions are not explicitly determined. Rather, inter- and intra-class variances are automatically combined into the optimal weighting matrix \allowbreak$\left({\mathbf{\Sigma}_B}^{-1} {\mathbf{\Sigma}_W}\right)^{1/2}$. Formulating a criterion similar to that in Equation~\eqref{e:pccc} now leads to eigenvalue decomposition of the matrix $\hat{\mathbf{\Phi}}_{XY}$:
\begin{align}
  \hat{\mathbf{\Phi}}_{XY} = \mathbf{\Upsilon}_X \mathbf{P}  \mathbf{\Upsilon}_Y
\end{align}
where:
\begin{align}
  \mathbf{P} = \left({\mathbf{\Sigma}_W}\right)^{1/2} \left({\mathbf{\Sigma}_B}\right)^{-1} \left({\mathbf{\Sigma}_W}\right)^{1/2}
\end{align}
Just as $\mathbf{\Phi}_{XY}$ in the non-discriminative case and using the same argument as on page~\pageref{p:phi}, $\hat{\mathbf{\Phi}}_{XY}$ can be recognized as a non-symmetric but nonetheless positive semi-definite matrix. Thus, the first discriminative extended canonical correlation coefficient is equal to its largest eigenvalue and, as in Equation~\eqref{e:pcca_sim}, the overall similarity of sets $\mathcal{X}$ and $\mathcal{Y}$ becomes:
\begin{align}
  \hat{\mu}_{XY} &=\frac { \sum_{i=1}^D \lambda_\phi^{(i)} } {\sum_{i=1}^D \sqrt{\lambda_X^{(i)} \lambda_Y^{(i)}}  }
    =\frac { \text{Tr}~\left[\hat{\mathbf{\Phi}}_{XY}\right] } {\sum_{i=1}^D \sqrt{\lambda_X^{(i)}~\lambda_Y^{(i)}}  }
\end{align}

\section{Experimental Evaluation}\label{s:eval}
We empirically examined the validity of theoretical arguments put forward in the preceding sections on the problem of matching sets of images of faces. To make our
experiments as directly comparable as possible to those in the published previous work most closely related to ours, we evaluated the proposed methods on a database already widely used for this purpose (e.g.\ see \cite{AranCipo2006e,KimAranCipo2007}) and described in detail in \cite{Aran2012}.

\begin{figure*}
  \centering
  \subfigure[Raw video frames]{\includegraphics[width=0.9\textwidth]{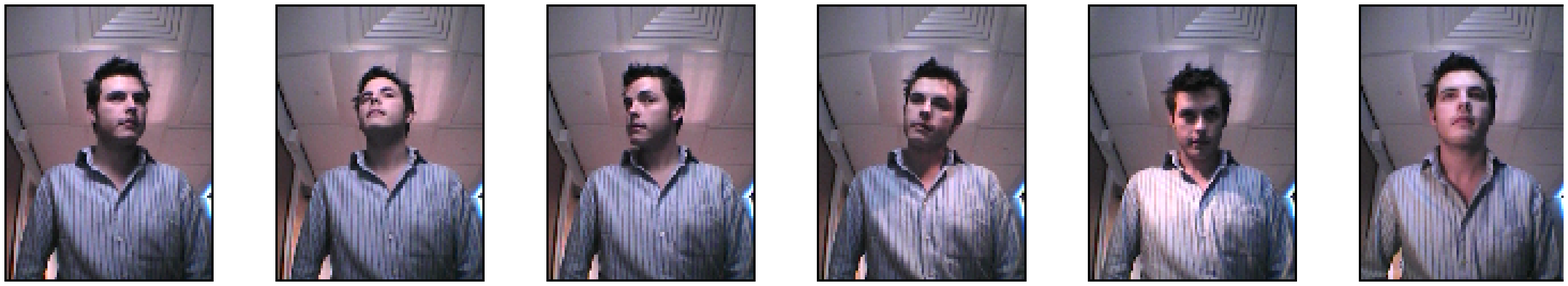}}
  \subfigure[Different illuminations]{\begin{tabular}{c}
    \includegraphics[height=0.13\textwidth]{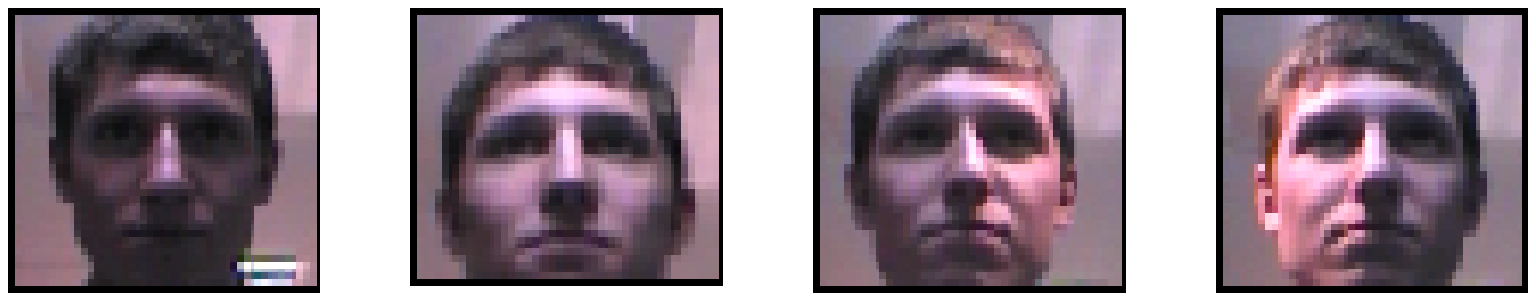}
    \includegraphics[height=0.13\textwidth]{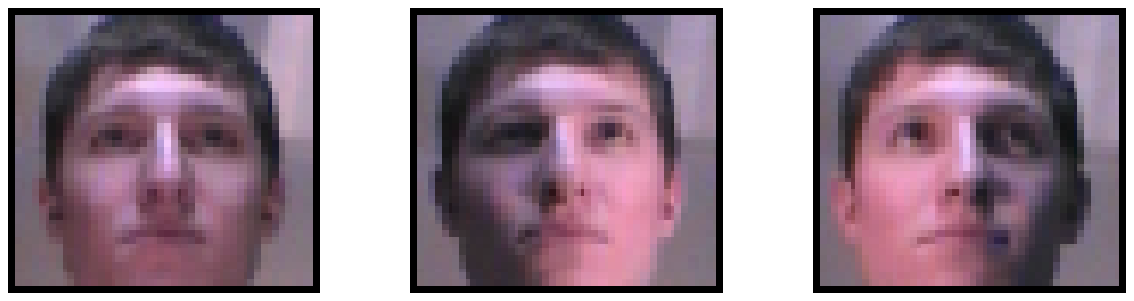}
  \end{tabular}}
  \vspace{10pt}
  \subfigure[Same illumination for different individuals]{\includegraphics[width=0.65\textwidth]{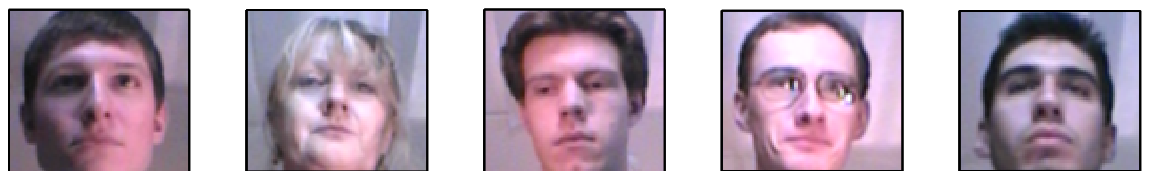}}
  \caption{ (a) Frames from typical video sequences in the database used for evaluation. (b) Different illumination conditions used to acquire data, illustrated on manually selected frontal faces. (c) Five different individuals in the illumination setting number~6. In spite of the same spatial arrangement of light sources, their effect on the appearance of faces changes significantly due to variations in people's heights, the \textit{ad lib} chosen position relative to the camera etc.  }
  \label{f:data}
\end{figure*}

This database contains video sequences of face motion for 100 individuals of varying ages and ethnicities. For each person in the database there are 7 sequences of the person performing loosely constrained, pseudo-random motion (significant translation, yaw and pitch, negligible roll) for 10~s, as shown on an example in Figure~\ref{f:data}(a). Each sequence was acquired in a different illumination setting as illustrated in Figure~\ref{f:data}(b). Sequences were acquired at 10~fps and in $320 \times 240$ pixel resolution (face size $\approx 60$ pixels)\footnote{A thorough description of the University of Cambridge face database with examples of video sequences is available at \texttt{http://mi.eng.cam.ac.uk/$\sim$oa214/}.}. The users were asked to approach the camera while performing arbitrary head motion. Although the illumination was kept constant throughout each sequence, there is some variation in the manner in which faces were lit due to the change in the relative position of the user with respect to the lighting sources, as shown in Figure~\ref{f:data}(c).

Faces were detected automatically using the cascaded detector of Viola and Jones \cite{ViolJone2004} are rescaled to the uniform resolution of $50 \times 50$ pixels which is approximately the average size of a detected face (see \cite{Aran2010b} for a related discussion). Face image patches were then converted into vectors by column-wise rasterization, each video sequence thus producing a set of vectors in the $\mathds{R}^{2500}$ space. Different distance measures between sets were evaluated in the context of one-to-many matching. In other words, image sets extracted from video sequences of different individuals in a particular setting were used as training data while querying was performed using sets corresponding to a different illumination. Each query set was associated with the best matching training set. In this manner, we investigated:
\begin{itemize}
  \item the sensitivity of the classical CCA to the number of correlation coefficients,\\
  \item the sensitivity of the C-CCA to the dimensionality of the constraining subspace,\\
  \item the performance of the proposed E-CCA distance measure, and\\
  \item the performance of the proposed DE-CCA distance measure.
\end{itemize}

\subsection{Results and Discussion}
\begin{figure}[!h]
  \centering
  \subfigure[Canonical correlation analysis]{\includegraphics[width=0.45\textwidth]{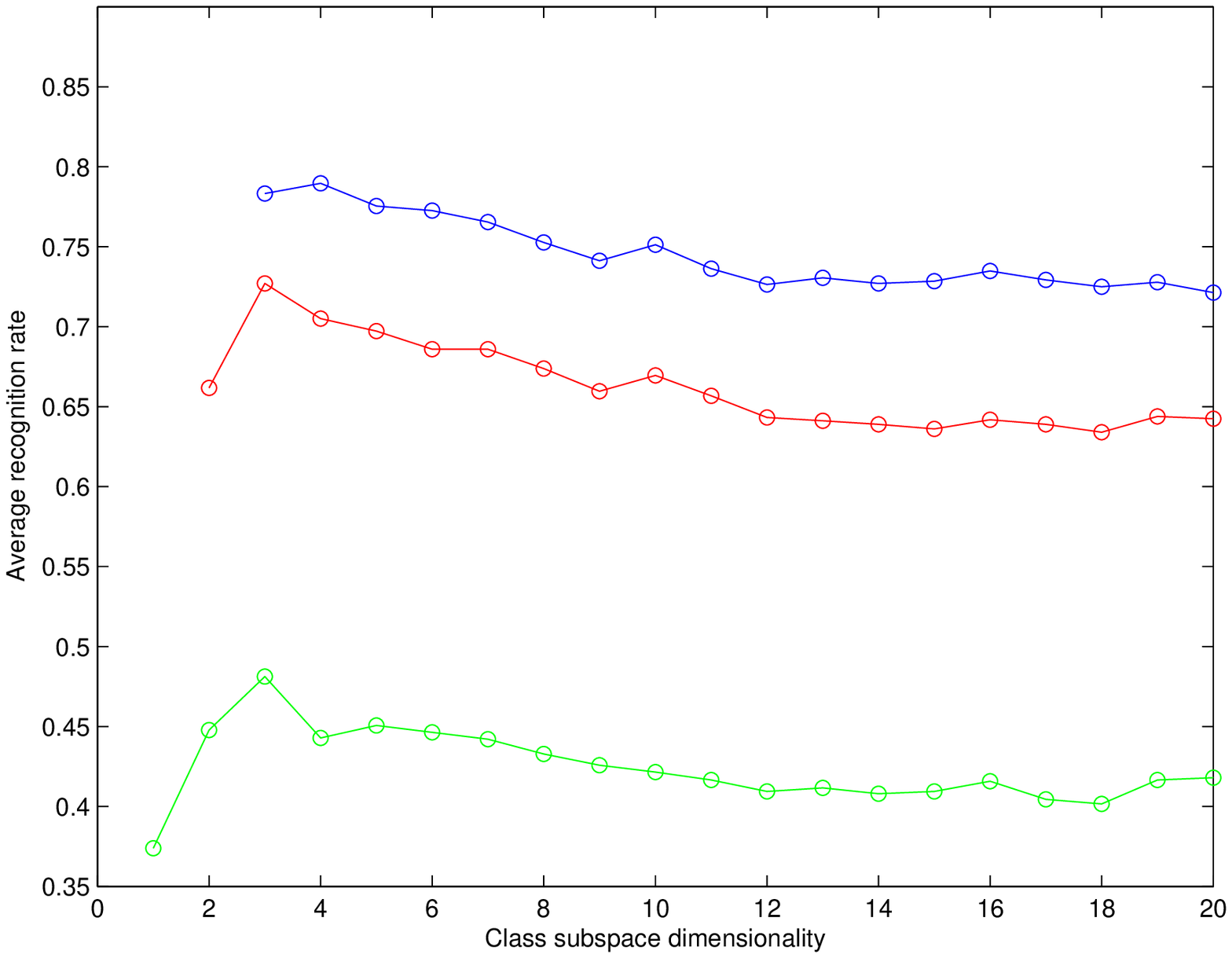}}
  \subfigure[Extended canonical correlation analysis]{\includegraphics[width=0.45\textwidth]{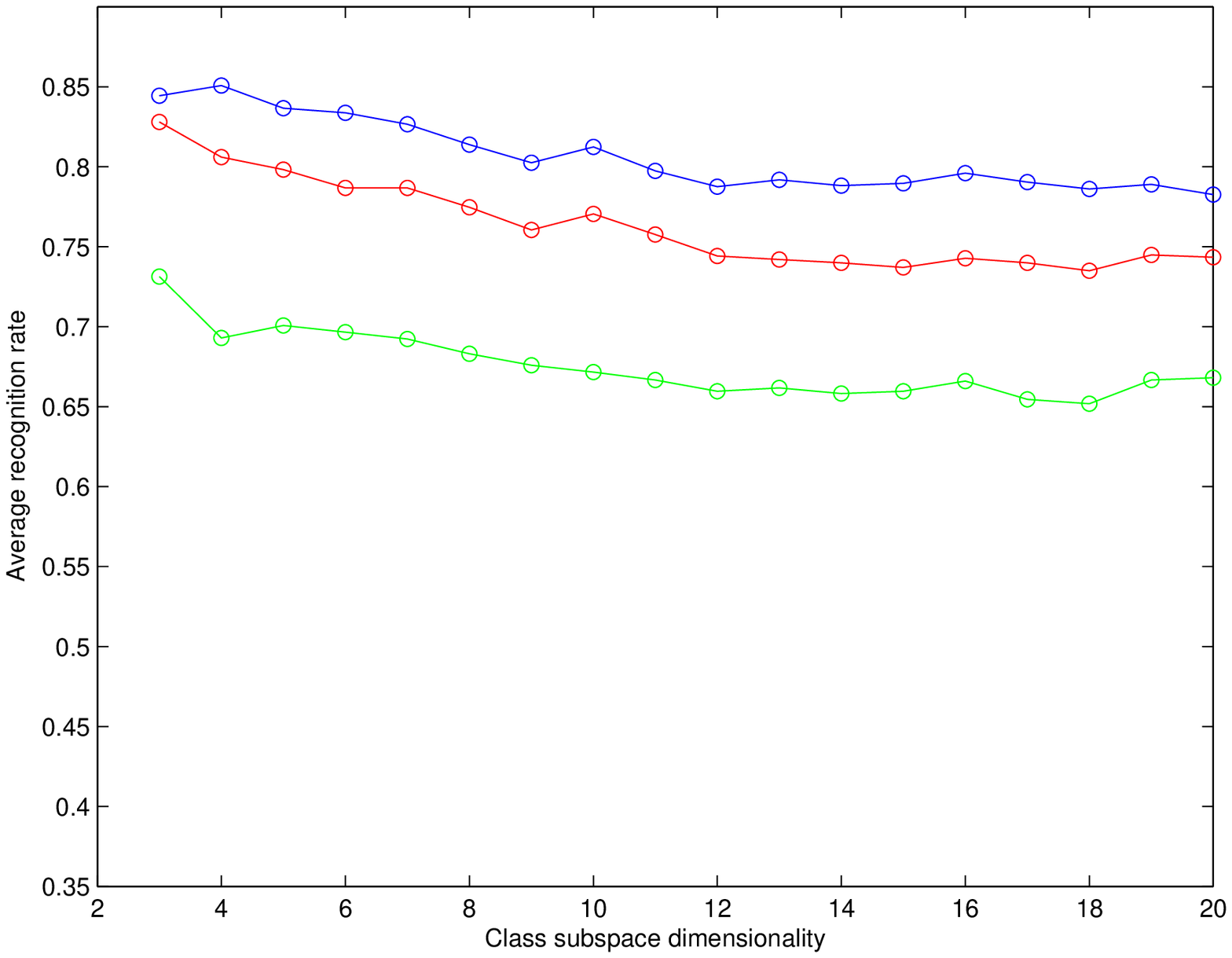}}
  \subfigure[Discriminative extended canonical correlation analysis]{\includegraphics[width=0.5\textwidth]{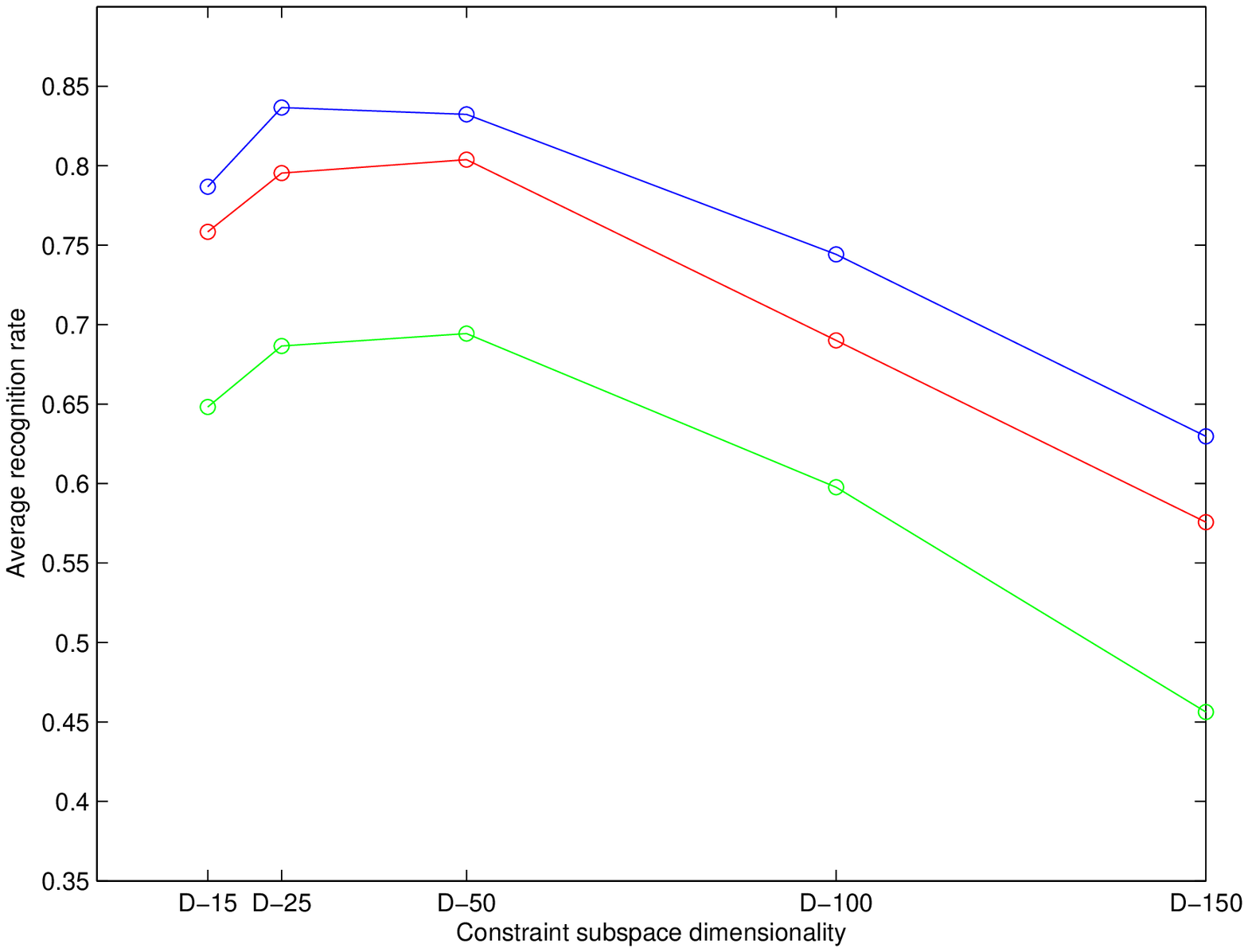}}
  \caption{ The measured variation of the correct recognition rate averaged over different illumination conditions used for training and querying, as a function
            of the dimensionality of the constraint subspace and the number of canonical correlation coefficients used for matching (green=1, red=2, blue=3). Optimal performance is achieved using three canonical correlations and $(D-25)$-dimensional constraint subspace.  }
  \label{f:dim}
\end{figure}

A summary of the key results is shown in Table~\ref{f:results}. As expected in the case
of data with such complex variability as exhibited by face appearance images, the performance of the
classical CCA-based matching is greatly affected by the number of canonical correlation
coefficients used to compute the set-to-set similarity measure. When their number is increased
from one to two, the average performance is improved by 59\%, and further by 78\% for
three coefficients. Additional sensitivity of the method to free parameter selection
can be observed in Figure~\ref{f:dim}(a), which illustrates the importance of the dimensionality
of class subspaces.

It is revealing to notice that while an increase in the number of computed
canonical correlations improves the average rate of correct identification, the confidence
in the recognition decision (quantified by
the deviation of the correct identification rate across different test and query illumination
conditions) actually worsens. This phenomenon can be explained by observing that increasing the
number of coefficients does not improve matching decisions in the most challenging cases
(corresponding to test and query illumination combinations that result in the lowest correct
identification rates). In other words, in these cases, the variations present in
different appearance sets corresponding to the same person are indeed very unlike one another,
and the only way of producing an improvement is by incorporating discriminative constraints.

\begin{table*}[tb]
  \centering
  \small
  \caption{ A summary of experimental results. Shown are the average correct recognition rate
            across different illuminations used for training and querying, and its
            standard deviation. Even for the optimal choice of parameters, which is impossible
            to ensure in practice, classical canonical correlation analysis (CCA: a non-discriminative
            approach) and its constrained extension (C-CCA: discriminative) were outperformed by
            the proposed methods which are parameter-free. }
  \begin{tabular}{|l|ccc|cccc|}
    \Hline
    \small ~Method             &  \multicolumn{3}{c|}{\small CCA}   & \multicolumn{4}{c|}{\small C-CCA (3 cc)} \vspace{-0pt}\\
           ~Parameters         & \small~~1 cc~ & \small~2 cc~ & \small~3 cc~~ & \small~~-15d~ & \small~-25d~ & \small~-50d~ & \small~-100d\\
    \hline
    \small~Mean correct recognition rate & \small 42.2 & \small 67.0 & \small 75.1 & \small 78.7 & \small 83.6 & \small 83.2 & \small 74.4\\
    \small~Recognition rate deviation~~  & \small 10.6  &  \small 13.9    &   \small 13.5   &  \small 11.1     &    \small 8.3     &  \small 9.0     & \small 15.4\\
    \Hline
  \end{tabular}\\
  \vspace{15pt}
  \begin{tabular}{|l|c|c|}
    \Hline
    \small ~Method                       &  \small ~~E-CCA~~ & \small ~~\textbf{DE-CCA}~~ \\
    \hline
    \small~Mean correct recognition rate &  \small 83.4      & \small \textbf{88.8}   \\
    \small~Recognition rate deviation~~  & \small 9.6        & \small \textbf{7.8}   \\
    \Hline
  \end{tabular}\\\vspace{15pt}
  \label{f:results}
\end{table*}

The results obtained using constrained CCA highlight further limitations of classical canonical correlation analysis in set matching. In addition to the sensitivity of the method to the free parameters shared with CCA -- the number of canonical correlations used to compute set similarities and the dimensionality of class subspaces, illustrated for C-CCA in Figure~\ref{f:dim}(b) -- an additional free parameter is introduced in the form of the dimensionality of the constraint subspace. Discriminative projection of data was found to improve performance only for constraint subspace dimensionalities greater than $(D-95)$, which is a narrow range of just $\sim$3.8\% of the input space dimensionality $D = 2500$. In this range, the mean recognition rate is increased and its deviation decreased. Peak correct recognition
rate of 83.6\% (33\% reduction in error rate compared to simple CCA) is achieved when three canonical correlation coefficients are used and the dimensionality of the constraint subspace set equal to $(D-25)$. However, a suboptimal choice of the constraint subspace dimensionality can rapidly lead to significant worsening of performance, as shown in Figure~\ref{f:dim}(c). This is a major limitation of C-CCA, as there is no fundamental theoretical basis which could facilitate
the inference of this parameter.

As this paper argued in detail, the problem of free parameter choice is made entirely obsolete by our use of second order statistics. The proposed E-CCA was found to significantly outperform not only CCA but C-CCA as well, and remarkably for \emph{all values of their free parameters}, as can be seen in Table~\ref{f:results}. Practically, this is a major advantage of our approach, due to the difficulty of ensuring optimal parameter selection for C-CCA. Theoretically, this confirms the premise set forth in Section~\ref{s:intro} which emphasizes the loss of discriminative information in discarding higher order statistics of class variability. This is an important result given that unlike E-CCA, C-CCA is a discriminative approach. With that in mind, it is not surprising that our discriminative method, DE-CCA, improves performance even further, decreasing the error rate by 32\% from C-CCA -- see Table~\ref{f:results}.

\subsection{Qualitative Insight}
\begin{figure}[tb]
  \centering
  \footnotesize
  \subfigure[E-CCA, same class]{\includegraphics[width=0.45\textwidth]{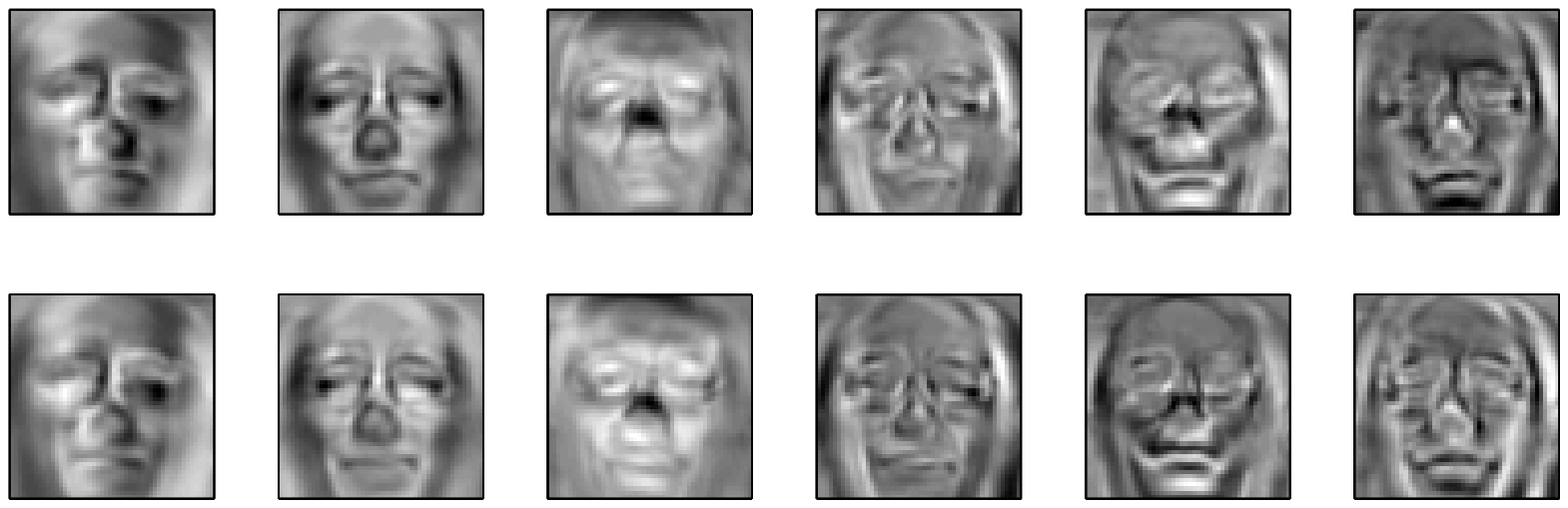}}\hspace{15pt}
  \subfigure[E-CCA, different classes]{\includegraphics[width=0.45\textwidth]{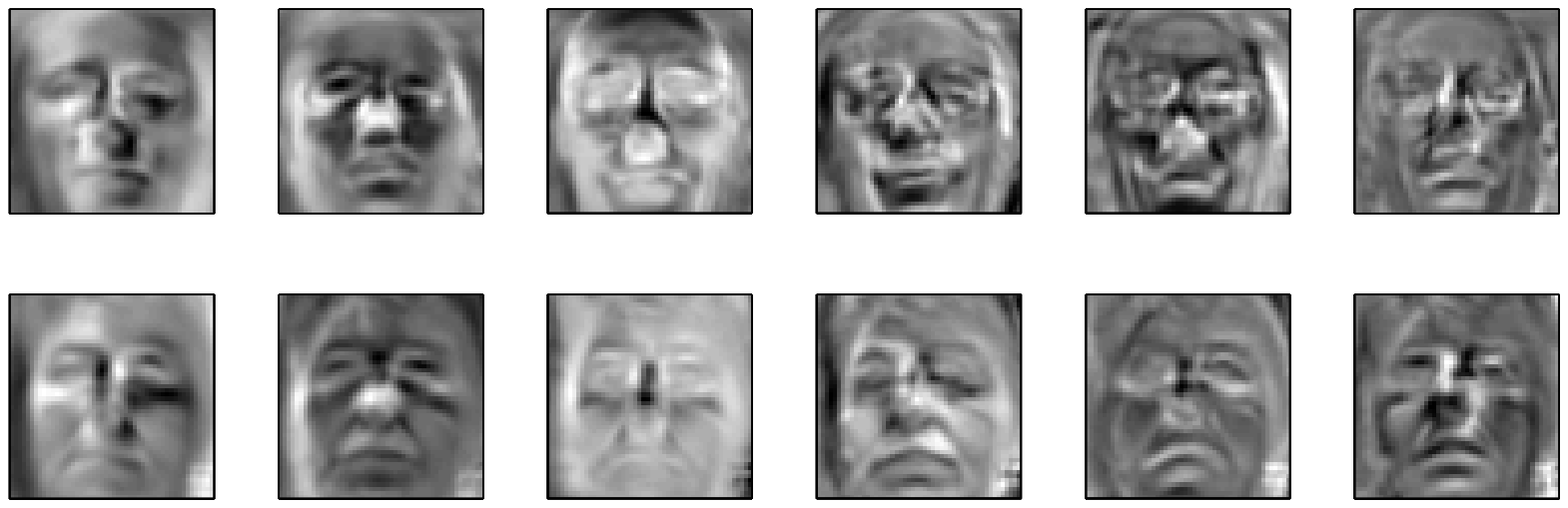}}
  \subfigure[DE-CCA, same class]{\includegraphics[width=0.45\textwidth]{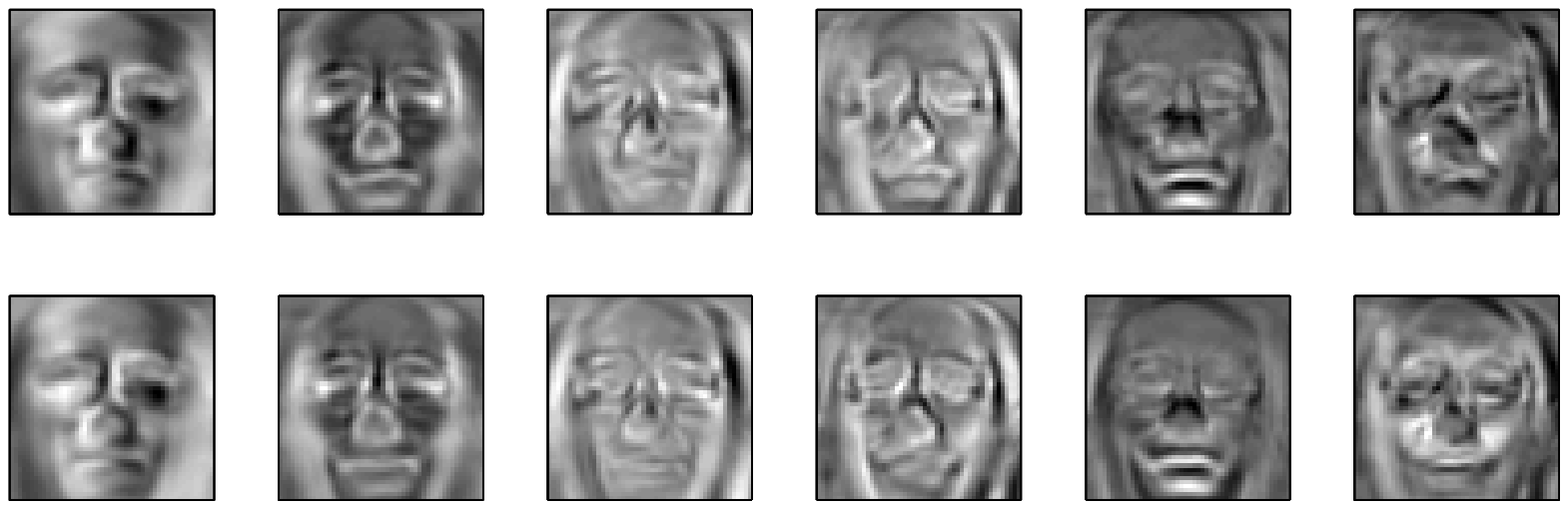}}\hspace{15pt}
  \subfigure[DE-CCA, different classes]{\includegraphics[width=0.45\textwidth]{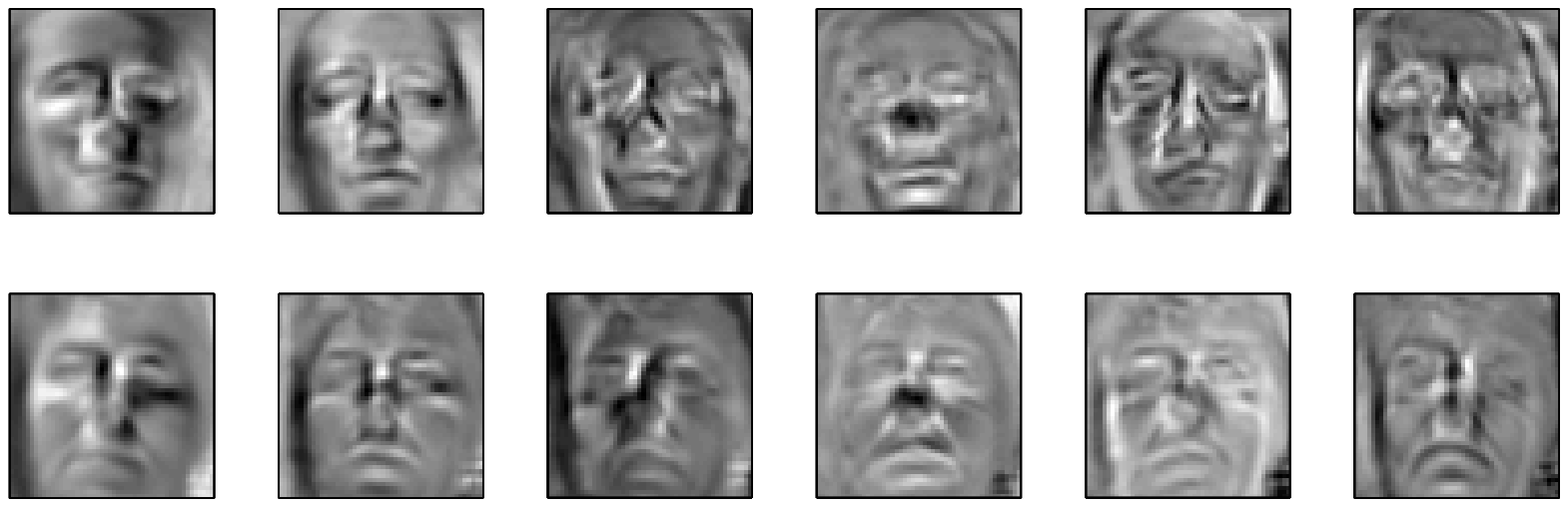}}
  \caption{ The most similar 6 modes of variation extracted using (a,b) E-CCA and (c,d) DE-CCA
            using sets corresponding to the same person in different illuminations (a,c) and
            different persons. In the case of same class matching, both methods successfully
            extract common directions of appearance change between two sets. The benefit of
            DE-CCA is more pronounced when sets of images of different persons are matched --
            E-CCA erroneously manages to find rather more alike modes of variation by effectively
            matching on common illumination, while the discriminative algorithm correctly
            learns and de-emphasizes such confounding factors. }
  \label{f:modes}
\end{figure}

It is insightful to visually examine the modes of the most probable common variability between sets of the same and different classes, extracted by the proposed
methods. These can be found as the dominant eigenvectors of the matrix $\mathbf{\Phi}_{XY}$
in the case of E-CCA (see Section~\ref{s:pcca}), and $\hat{\mathbf{\Phi}}_{XY}$ (see
Section~\ref{s:dpcca}) in the case of DE-CCA. Examples, visualized as images, are shown in
Figure~\ref{f:modes}. Notice that E-CCA manages to find meaningful
common modes even when the two sets correspond to different people, just as it does
when they show the same person -- it effectively matches similar appearing illumination
effects. This is a consequence of geometric and textural similarity of human faces, which makes
face recognition so difficult. By learning a discriminative criterion, matching
using DE-CCA has the consequence of de-emphasizing confounding inter-personal variability and
amplifying intra-personal information.

\subsection{Summary and Conclusions}\label{s:conc}
In this paper we proposed a novel framework for matching vector sets. Our approach is based on inference
of the most similar modes of variability within two sets. This led to a comparison with increasingly popular canonical
correlation analysis-based methods. These were discussed in detail and it was shown, both theoretically and
empirically, that they have significant practical limitations of which the most important are: (i) the presence
of free parameters which cannot be inferred from the data, (ii) non-robust partitioning of the space into class
and non-class subspaces,
and (iii) intractability of discriminative learning. In contrast, the proposed \emph{extended canonical
correlation analysis}-based method inherently accounts for
uncertainty in data, inferring \emph{the most likely} common modes of variability.  What is more, it is shown
that this can be achieved without the loss of attractive computational efficiency of canonical correlation
analysis, whereby set similarity is effectively reduced to the computation of the trace of a matrix. The
proposed framework was then extended into a discriminative learning scheme which, unlike in the case of classical
canonical correlation analysis, follows naturally. Finally, our theoretical arguments were empirically
verified on the task of set-based face recognition. The proposed methods were shown superior even for the
optimal choice of parameters of the  classical canonical correlation analysis-based methods, which is impossible
to ensure in practice.

\bibliographystyle{unsrt}
\bibliography{./my_bibliography}

\end{document}